\documentclass[nonatbib]{article}

\usepackage[final]{neurips_2023}

\usepackage[utf8]{inputenc} %
\usepackage[T1]{fontenc}    %
\usepackage{hyperref}       %
\usepackage{url}            %
\usepackage{booktabs}       %
\usepackage{csquotes}
\usepackage{amsfonts}       %
\usepackage{nicefrac}       %
\usepackage{microtype}      %

\usepackage{float}

\usepackage{adjustbox}      %
\usepackage[sorting=nyt,style=authoryear,bibencoding=utf8,backend=biber,natbib=true,uniquename=false,uniquelist=false,maxbibnames=99]{biblatex}
\usepackage[table,x11names,dvipsnames,table]{xcolor}

\title{How to use and interpret activation patching}

\author{%
  Stefan Heimersheim \\
  \texttt{stefan.heimersheim@gmail.com} \\
  \And
  Neel Nanda \\
}

\begin{document}

\maketitle

\begin{abstract}
  Activation patching is a popular mechanistic interpretability technique, but has many subtleties regarding how it is applied and how one may interpret the results. We provide a summary of advice and best practices, based on our experience using this technique in practice. We include an overview of the different ways to apply activation patching and a discussion on how to interpret the results. We focus on what evidence patching experiments provide about circuits, and on the choice of metric and associated pitfalls.
\end{abstract}

\section{Introduction}
\subsection{What is activation patching?}
Activation patching (also referred to as Interchange Intervention, Causal Tracing, Resample Ablation, or Causal Mediation Analysis) is the technique of replacing internal activations of a neural net. 
It is also known as Causal Tracing, Resample Ablation, Interchange Intervention or more generally as Causal Mediation Analysis. Variants of this technique have been widely used in the literature \citep{2020arXiv200412265V,2021arXiv210602997G,2020arXiv200414623G,2019arXiv191009113S,2021arXiv210606087F,2021arXiv211200826G,ROME,IOI,causal_scrubbing,2023arXiv230104213H,2023arXiv230500586H,ACDC,2023arXiv231015213T,2023arXiv231015916H,2023arXiv231017191F,LieberumChincilla,2023arXiv230908600C,2023arXiv230515054S,NixPathPatching,CallumCopySuppression,2023arXiv230414767G,2023arXiv230910312H,2023arXiv231008744M,2023arXiv231015154T,}. Here we focus on the technique where we overwrite some activations during a model run with cached activations from a previous run (on a different input), and observe how this affects the model’s output.

\subsection{How is this related to ablation?}
Ablation is the common technique of zeroing out activations. Activation patching is more targeted and controlled: We replace activations with other activations rather than zeroing them out. This allows us to make targeted manipulations to locate specific model behaviours and circuits.

\subsection{An example}
For example, let’s say we want to know which model internals are responsible for factual recall in ROME \citep{ROME}. How does the model complete the prompt “The Colosseum is in” with the answer “Rome”? To answer this question we want to manipulate the model’s activations. But the model activations contain many bits of information: This is an English sentence; The landmark in question is the Colosseum; This is a factual statement about a location.

Ablating some activations will affect the model if these activations are relevant for \textit{any} of these bits. But activation patching allows us to \textit{choose} which bit to change and \textit{control} for the others. Patching with activations from “Il Colosseo è dentro” locates where the model stores the language of the prompt but may use the same factual recall machinery. Patching with activations from “The Louvre is in” locates which part of the model deals with the landmark and information recall. Patching between “The Colosseum is in the city of” and “The Colosseum is in the country of” locates the part of the model that determines \textit{which} attributes of an entity are recalled. 

A simple activation patching procedure typically looks like this:
\begin{enumerate}
  \item Choose two similar prompts that differ in some key fact or otherwise elicit different model behaviour:
  \begin{itemize}
    \item[] E.g. “The Colosseum is in” and “The Louvre is in” to vary the landmark but control for everything else.
  \end{itemize}
  \item Choose which model activations to patch
  \begin{itemize}
    \item[] E.g. MLP outputs
  \end{itemize}
  \item Run the model with the first prompt—the source prompt—and save its internal activations
  \begin{itemize}
    \item[] E.g. “The Louvre is in” (source)
  \end{itemize}
  \item Run the model with the second prompt—the destination prompt—but overwrite the selected internal activations with the previously saved ones (patching)
  \begin{itemize}
    \item[] E.g. “The Colosseum is in” (destination)
  \end{itemize}
  \item See how the model output has changed. The outputs of this patched run are typically somewhere between what the model would output for the un-patched first prompt or second prompt
  \begin{itemize}
    \item[] E.g. observe change in the output logits for "Paris" and "Rome"
  \end{itemize}
  \item Repeat for all activations of interest
  \begin{itemize}
    \item[] E.g. sweep to test all MLP layers
  \end{itemize}
\end{enumerate}

\subsection{What is this document about}
We want to communicate useful practical advice for activation patching, and warn of common pitfalls to avoid. We focus on three areas in particular:
\begin{enumerate}
    \item What kind of patching experiments provide which evidence? (Section \ref{sec1})
    \item How should you interpret activation patching results? (Section \ref{sec2})
    \item What metrics you can use, what are common pitfalls? (Section \ref{sec3})
\end{enumerate}

For a general introduction to mechanistic interpretability in general, and activation patching in particular we refer to ARENA\footnote{ARENA: \url{https://arena3-chapter1-transformer-interp.streamlit.app/}} chapter 1 (in particular activation patching in chapter 1.3) as well as the corresponding glossary entries on Neel Nanda’s website\footnote{\url{https://neelnanda.io/glossary}}.

\section{What kind of patching experiments should you run?}
\label{sec1}
\subsection{Exploratory and confirmatory experiments}
In practice we tend to find ourselves in one of two different modes of operation: In exploratory mode we run experiments to find circuits and generate hypotheses. In confirmatory mode we want to verify the circuit we found and check if our hypothesis about its function is correct.

In \textit{exploratory} patching we typically patch components one at a time, often in a sweep over the model (layers, positions, model components). We do this to get an idea of which parts of a model are involved in the task in question, and may be part of the corresponding circuit.

In \textit{confirmatory} patching we want to confirm a hypothesised circuit by verifying that it actually covers all model components needed to perform the task in question. We typically do this by patching many model components at once and checking whether the task performance behaves as expected. A well-known example of patching for circuit verification is Causal Scrubbing \citep{causal_scrubbing}.

\subsection{Which components should you patch}
Patching can be done on different levels of granularity determining the components to patch. For example, we may patch the residual stream at a certain \textit{layer} and \textit{position}, or the output of a certain MLP [\textit{layer}, \textit{position}] or Attention Head [\textit{layer}, \textit{head}, \textit{position}]. At even higher granularity we could patch individual neurons or sparse autoencoder features.

An even more specific type of patching is path patching. Usually, patching any component will affect \textit{all} model components in later layers (“downstream”). In path patching instead we let each patch affect only a single target component. We call this patching the “path” between two components. For details on patch patching we refer to ARENA section 1.3.4.

Path Patching can be used to understand whether circuit components affect each other directly, or via mediation by another component. For example if we want to distinguish between \textit{mediation} (component A affects output C via component B), and \textit{amplification/calibration} (component A affects output C directly, but component B reads from A and also affects output C by boosting or cancelling the signal to amplify or calibrate component A). These two structures look identical in default component patching, but different in path patching: a direct connection (composition) between A and C exists only in the second case.

As a rule of thumb, you want to start with low-granularity patching (e.g. residual stream patching), then increase granularity, and finally use path patching to test which components interact with each other. Fast approximations to activation patching, such as attribution patching (see \cite{attribution_patching}, and also
AtP*, \cite{atpstar}) can help speed up this process in large models.

\subsection{Noising and Denoising}

There are multiple ways to do activation patching. The techniques differ in what the source (source of activations / model run from which the activations are copied) and destination prompt (destination that is overwritten / model run in which the activations will be inserted, this is called base in Interchange Interventions language, \cite{2021arXiv211200826G}) are. \textit{The use of words “source” and “destination” is unrelated to their meaning in Transformer attention.}

The two main methods are Denoising and Noising (see the next section for other methods).

\begin{enumerate}
    \item \textbf{Denoising}: We can patch activations from a clean first prompt into a corrupted second prompt “clean → corrupt”. That is running the model on the clean prompt while saving its activations, then running the model on the corrupted prompt while overwriting some of its activations with previously saved clean-prompt activations. We observe which patch restores the clean-prompt behaviour, i.e. patching which activations were \textbf{sufficient to restore} the behaviour.
    \item \textbf{Noising}: Or you can patch activations from a corrupted first prompt into a clean second prompt “corrupt → clean”. That is running the model on the corrupted prompt while saving its activations, then running the model on the clean prompt while overwriting some of its activations with previously saved corrupt-prompt activations. We observe which patch breaks the clean-prompt behaviour, i.e. patching which activations were \textbf{necessary to maintain} for the behaviour.
\end{enumerate}

An important and underrated point is that these two directions can be very different, and are not just symmetric mirrors of each other. In some situations denoising is the right tool, and in others it’s noising, and understanding the differences is a crucial step in using patching correctly.

\begin{adjustbox}{width=1.2\textwidth,center=\textwidth}
  \begin{tabular}{p{3.6cm}p{2.5cm}lp{3.5cm}p{2.3cm}p{1.89cm}}
    \toprule
    Technique & Source (saved) & Source run input & Destination / Base (overwritten) & Destination / Base run input & Observation \\
    \midrule
    Clean $\rightarrow$ corrupted (Denoising, Causal Tracing\footnotemark[2]) & First run activations (clean) & Clean tokens & Second run activations (corrupted) & Corrupt tokens & What restores behaviour \\
    Corrupted $\rightarrow$ clean (Noising, Resample Ablation) & First run activations (corrupted) & Corrupt tokens & Second run activations (clean) & Clean tokens & What breaks behaviour \\
    \bottomrule
  \end{tabular}
\end{adjustbox}

For now we round patching effects to “if I patch these activations the model performance is / isn’t affected”. We discuss metrics and measuring patching effects in the last section.

\subsection{Example: AND gate vs OR gate}
Consider a hypothetical circuit of three components A, B, and C that are connected with an AND or an OR gate. They are embedded in a much larger network, and of the three just C is connected to the output. We run an experiment where we patch all components using the denoising or noising technique.

\paragraph{AND circuit:} C = A AND B
\begin{itemize}
    \item Denoising (clean → corrupt patching): Denoising either A or B has no effect on the output, only denoising C restores the output. This is because denoising A still leaves B at the corrupted (incorrect) baseline, and vice versa. \textit{Denoising found only one of the circuit components.}
    \item Noising (corrupt → clean patching): Noising either A or B has an effect, as well as noising C.
\end{itemize}

\textit{Noising works better in this case, as it finds all circuit components in the first pass.
}
\paragraph{OR circuit:} C = A OR B
\begin{itemize}
    \item Denoising (clean → corrupt patching): Denoising either A or B has an effect, as well as denoising C.
    \item Noising (corrupt → clean patching): Noising either A or B has no effect on the output, only denoising C restores the output. This is because noising A still leaves B at the clean (correct) baseline, and vice versa. \textit{Denoising found only one of the circuit components.}
\end{itemize}

\textit{Denoising works better in this case, as it finds all circuit components in the first pass.
}
These AND and OR structures can appear in real-world transformers as serial-dependent components (e.g. a later attention head depending on an earlier one) or parallel components (such as redundant backup attention heads).

\subsection{Comparison to ablations and other patching techniques}
There are activation patching techniques based on a single prompt. The original Causal Tracing \citep[ROME,][]{ROME} falls into this category, and also zero- and mean-ablation can be seen as patching techniques.
\begin{enumerate}
    \item \textbf{Zero ablation}: Overwrite (“ablate”) the targeted activations with zeros and observe ablating which component breaks the model behaviour.
    \item \textbf{Mean ablation}: Same as above but overwrite targeted activations with their dataset mean value rather than zero. This is slightly more principled than zero ablating since there is no special meaning to activations being zero.
    \item \textbf{Gaussian noise patching} (also called Causal Tracing*): This is a clean → corrupt patching variant that uses as its corrupt run input the embeddings of the clean prompt with added Gaussian noise. The idea is to thereby automatically generate the corrupted “prompt”. It was originally used in ROME (called Causal Tracing there) but has not been used much recently, especially because the corruption can sometimes be ineffective.\footnote{The success of Gaussian noise corruption is highly sensitive to the noise level. \citet{Zhang2023} that if the noise level is just slightly lower than used in ROME, the model can recover the correct completion despite the corruption.}
\end{enumerate}

\begin{adjustbox}{width=1.2\textwidth,center=\textwidth}
  \rowcolors{2}{white}{gray!25} %
  \begin{tabular}{p{3.6cm}p{2.5cm}lp{3.5cm}p{2.3cm}p{1.89cm}}
    \toprule
    Technique & Source (saved) & Source run input & Destination / Base (overwritten) & Destination / Base run input & Observation \\
    \midrule
    Clean $\rightarrow$ corrupted (Denoising, Causal Tracing\footnotemark[2]) & First run activations (clean) & Clean tokens & Second run activations (corrupted) & Corrupt tokens & What restores behaviour \\
    Corrupted $\rightarrow$ clean (Noising, Resample Ablation) & First run activations (corrupted) & Corrupt tokens & Second run activations (clean) & Clean tokens & What breaks behaviour \\
    Zero ablation & Zero activations & N/A & Clean run activations & Clean tokens & What breaks behaviour \\
    Mean ablation & Dataset mean activations & N/A & Clean run activations & Clean tokens & What breaks behaviour \\
    Gaussian Noise patching (Causal Tracing\footnotemark[2]) & First run activations (clean) & Clean tokens & Second run activations (corrupted from modified clean input) & Clean token embedding + Gaussian noise & What restores behaviour \\
    \bottomrule
  \end{tabular}
\end{adjustbox}

\footnotetext[2]{Causal Tracing has been used to describe ROME-style Gaussian noise patching in particular, but also to describe clean → corrupted patching in general. We recommend avoiding the name to avoid confusion.}

Generally we recommend corrupted-prompt-based techniques, noising and denoising. Their advantage is that one can run very precise experiments, editing some features while controlling for others. They allow us to trace \textit{the difference between clean and corrupted prompt}. To illustrate this consider the prompts “Angela Merkel is the leader of” → “Germany” vs “Joe Biden is the leader of” → “America”. Patching will find components that deal with Angela Merkel vs Joe Biden, but not components that would be indifferent to this change, such as the “answer is a country circuit” or the “political leader circuit”. A secondary advantage of noising and denoising is that they tend to bring the model less out-of-distribution than ablation techniques (as pointed out in \citet{causal_scrubbing}, as well as in e.g. \cite{2021arXiv210600786H})

\subsection{Choosing corrupted prompts}
Having a corrupted prompt is great because it can tell us what model components care about, but also a possible pitfall if we don’t notice what our prompts trace and don’t trace. We give some examples for the Indirect Object Identification \citep[IOI,][]{IOI} demo sentence “John and Mary went to the store. John gave a bottle of milk to”. Different corruptions which highlight different properties the model might care about include:

\begin{adjustbox}{width=1\textwidth,center=\textwidth}
  \rowcolors{2}{white}{gray!15}
  \begin{tabular}{p{6.6cm}p{4cm}p{3.5cm}}
    \toprule
    Corruption & Example & Property traced in model \\
    \midrule
    None (Clean) & John and Mary … John … & \\
    Replace the value of one or multiple names, without changing the grammatical structure &
      John and Alice … John …
      Alice and Mary … Alice …
      Alice and Bob … Alice … &
      Where the model represents the name values \\
    Change which name is direct and indirect object without changing the names or positions &
      John and Mary … Mary … &
      The value and position of the indirect object \\
    Change the position of the names without changing which one is subject and indirect object &
      Mary and John … Mary … &
      The value, but not the position of the indirect object (position is fixed)  \\
    Change a name to break the behaviour &
      John and Mary … Alice …
      Alice and Mary … John … &
      Specifics about IOI setting (e.g. that a name is duplicated at all) \\
    Change all the names &
      Alice and Bob … Charlie … &
      Finding and confirming all relevant components \\
    \bottomrule
  \end{tabular}
\end{adjustbox}
\vspace{0.15cm}

What kind of prompt should you choose? No matter which you choose, keep in mind what properties your prompt does and does not change, and take this into account when interpreting patching results. As a rule of thumb you want to choose small (narrow) variations for exploratory patching, this will help you narrow down what each component is tracking. Choosing a narrow prompt distribution also helps increase the (typically low) sensitivity of denoising, and decrease the (typically high) sensitivity of noising. For confirmatory patching you need to choose a wide distribution of prompts that varies all variables of the hypothesised circuit. Then you can noise (corrupt → clean patch) all non-circuit components, and check that the model still shows the behaviour in question.

\section{How do you interpret patching evidence?}
\label{sec2}
In the previous section we said that denoising (clean → corrupt patching) tests whether the patched activations are \textbf{sufficient to restore} model behaviour. And noising (corrupt → clean patching)  tests whether the patched activations are \textbf{necessary to maintain} model behaviour. These two are usually not complements of each other, nor does one imply the other. In this section we will walk through a made-up example experiment.

\subsection{Walkthrough a stylized example}
Consider the hypothetical “Nobel Peace Prize” circuit. The model correctly completes “Nobel Peace” with “Prize”, using the following circuit:
\begin{itemize}
    \item Attention head L0H0 is a “Previous Token Head” and copies the embedding of  “Nobel” to the position of “Peace”
    \item Neuron L1N42 maps the mix of Nobel and Peace embeddings to the Prize logit
    \item Everything else doesn’t matter (of course a real circuit is typically much messier)
\end{itemize}
\begin{figure}[H]
    \centering
    \includegraphics[width=0.5\textwidth]{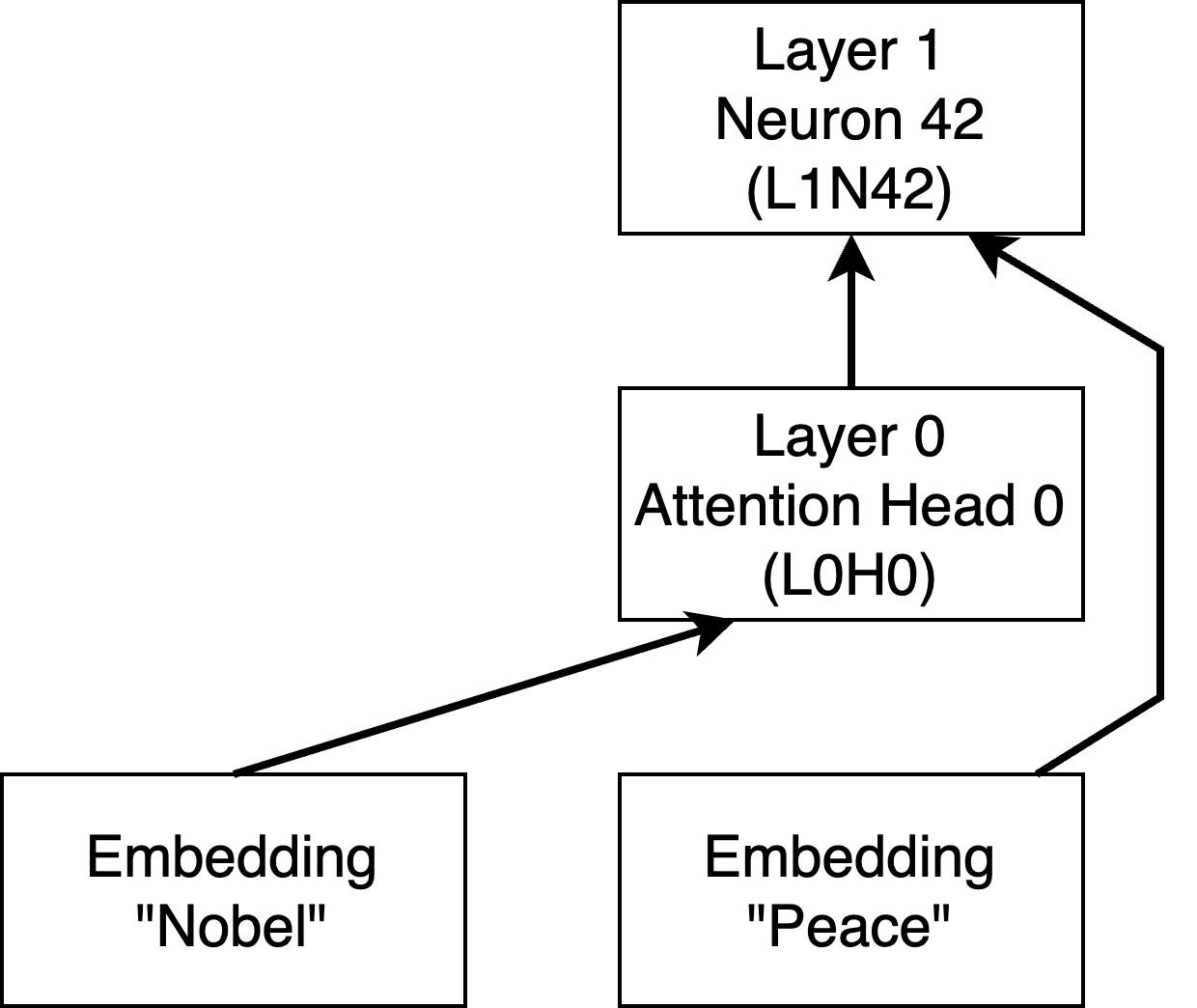}
    \caption{Toy "Nobel Peace Price" circuit}
    \label{nobelcircuit}
\end{figure}

Now let us run the standard patching examples, take a distribution of random English words for the corrupted prompt. We would find
\begin{itemize}
    \item Noising (corrupt → clean patching) suggests that the outputs of head L0H0, the output of neuron L1N42, and the embeddings (Nobel \& Peace) are all necessary components.
    \item Denoising (clean → corrupt patching) suggests that the output of neuron L1N42 is sufficient to restore the circuit.
\end{itemize}

What happened here? Denoising finds only the neuron output  L1N42, because the other two components individually are not sufficient to restore the circuit behaviour! We’re dealing with an AND circuit between the attention head output and the “Peace” embedding. Noising finds all three components here.

Nonetheless denoising L1N42 alone restored the model behaviour. This is a crucial intuition to keep in mind about denoising: If you patch component A in layer N, it has seen clean versions of every component in layers 0 to N-1. If there’s an important component B in layer N-1 that is mediated by component A, the model can be restored without denoising B.

Patching experiments are sensitive to what precisely are the changes between the corrupt and clean prompt. If we created two additional corrupt distributions where we replace only either “Nobel” or “Peace” with a random word (i.e. distributions “X Peace” and “Nobel Y”) we could narrow down which component depends on which input.

Alternatively we could use path patching to confirm the precise interactions. Say we want to test whether the Peace embedding is necessary as an input to L0H0, as an input to L1N42, or both. For this we could patch only the corresponding paths, and find that denoising (1) “Nobel → L0H0” and (2) “Peace → L1N42” paths is sufficient. Alternatively we might find that noising every path except for (1) “Nobel → L0H0”, (2) “L0 → L1N42”, and (3) “Peace → L1N42” does not break performance. Note again that denoising only required restoring two paths (restoring a cross-section of the circuit) while noising required leaving 3 paths clean (the full circuit).\footnote{This method doesn’t yet confirm which information is carried in the different paths. We can go a step further and noise (corrupt → clean patch) even some of the important circuit connections, namely “Nobel → L0H0 → L1N42” path from the “Nobel Y” distribution, and the “Peace → L1N42” path from the “X Peace” distribution. Doing that is essentially Causal Scrubbing \citep{causal_scrubbing}.}

\subsection{Concepts \& gotchas}
The walkthrough above presents a typical circuit discovery workflow. We want to highlight a couple of additional concepts and common issues.

\paragraph{Sensitivity \& prompt choice:} A positive patching result implies you have found activations dealing with the \textit{difference between the clean and corrupt prompt}. Make sure to consider all degrees of freedom in a task, and consider multiple sets of corrupted prompts if necessary.

\paragraph{Scope of activation patching:} More generally, activation patching is always based on prompt distributions, and does not make statements for model behaviour outside these specific distributions. For more discussion on the limitations of patching, and the specificity of prompt-based interpretability in general, see Neel Nanda’s writing on
\href{https://www.neelnanda.io/mechanistic-interpretability/attribution-patching#what-cant-activation-patching-teach-us=}{What Can(’t) Activation Patching Teach Us.}

\paragraph{No minimality:} Here, and in many parts of the literature, a circuit is treated as a collection of model components that are responsible for a particular model behaviour. We typically make no claims that we have found the \textit{smallest} such collection of components, we only test that this collection is sufficient.

\paragraph{Backup behaviour \& OR-gates:} In some cases researchers have discovered “Backup heads”, components that are not normally doing the task but jump into action of other components are disrupted 
\citep[Hydra effect,][]{2023arXiv230715771M}.
For example, in IOI when one ablates a name mover head (a key component of the circuit) a backup name mover head will activate and then do the task instead \citep{IOI}.

It can be helpful to think of these as OR-gates where either component is sufficient for the model to work. This does not fit well into our attempts of defining a circuit, nor plays well with the circuit finding methods above. Despite the name mover heads being important, if we ablate them then, due to backup heads compensating, the name movers look less important. Fortunately, backup behaviour seems to be lossy, i.e. if the original component boosted the logits by +X, the backup compensates for this by boosting less than X (the Hydra effect paper found 0.7*X). Thus these backup component weaken the visibility of the original component, but it is usually still visible since even 0.3*X is a relatively large effect.

\paragraph{Negative components:} Some work in this area \citep[e.g.][]{IOI,docstrings} noticed attention heads that consistently negatively affected performance, and noising them would increase performance. This is problematic, because it makes it hard to judge the quality of a circuit analysis: it may look like we’ve fully recovered (or more than fully recovered!) performance, by finding half the positive components but excluding all negative ones. This is an unsolved problem. \citet{ACDC} propose using Kullback Leibler (KL) divergence as a metric to address this, which penalises any deviation (positive or negative), at the cost of also tracking lots of variation we may not care about.

\section{Metrics and common pitfalls}
\label{sec3}
So far we talked about “preserving” and “restoring” performance, but in practice, model performance is not binary but a scale. Typically we find some components matter a lot, while others provide a small increase in performance. For the best interpretability we might look for a circuit restoring e.g. 90\% of the model’s performance, rather than reaching exactly 100\% \citep[for examples see][]{causal_scrubbing}. A useful framing is the “pareto frontier” of circuit size vs. performance recovered - recovering 80\% of performance with 1\% of the components is more impressive than 90\% of the performance with 10\% of the components, but there will always be a minimum circuit size to recover a given level of performance.

It’s easy to treat metrics as an after-thought, but we believe that the right or wrong choice of a metric can significantly change the interpretation of patching results. Especially for exploratory patching, the wrong metric can be misleading. The choice of metric matters less for confirmatory patching, where you expect a binary-ish answer (“have I found the circuit or not”) and all metrics should agree. We’ll go through a couple of metric choices in this section:

\begin{table}[h!]
  \begin{tabular}{ll}
    \toprule
    Based on &
    Example \\
    \midrule
    Logit difference (= Logprob difference) & Logit(Mary) - Logit(John) \\
    Logarithmic probability (logsoftmax) & Logprob(Mary) \\
    Probability (softmax) & Prob(Mary) \\
    Accuracy / Rank of correct answer & Rank(Mary)==0 \\
    \bottomrule
  \end{tabular}
\end{table}

An honourable mention goes to the KL divergence. Unlike the previous metrics, this metric aims to compare the full model output, rather than focusing on a specific task. KL divergence is a good metric in such cases.

In addition to these output based metrics, in some cases it makes sense to consider some model internals as metrics themselves. For example, one might use the attention paid by the name mover head to the indirect object as a metric to identify the subcircuit controlling this head, or the activation of a key neuron or SAE feature, or the projection onto a probe \citep{fact_finding}.

In our experience, it’s worth implementing many metrics and briefly analysing all of them. Computing a metric is cheap (compared to the cost of the forward pass), and they all have different strengths and weaknesses, and can illuminate different parts of the big picture. And if they all agree that’s stronger evidence than any metric on its own. Where they disagree, we personally trust \textbf{logit difference} (or equivalently logprob difference) the most.

\subsection{The logit difference}
Logit difference measures to what extent the model knows the correct answer, and it allows us to be specific: We can control for things we don’t want to measure (e.g. components that boost both, Mary and John, in the IOI example) by choosing the right logits to compare (e.g. Mary vs John, or multiple-choice answers). The metric also is a mostly linear function of the residual stream (unlike probability-based metrics) which makes it easy to directly attribute logit difference to individual components (“direct logit attribution”, “logit lens”). It’s also a “softer” metric, allowing us to see partial effects on the model even if they don’t change the rank of the output tokens (unlike e.g. accuracy), which is crucial for exploratory patching. We discuss problems with this and other metrics in the next section.

\textit{Intuition for why logits and logit differences (LDs) are a natural unit for transformers:} The residual stream and output of a transformer is a sum of components. Every component added to the residual stream corresponds to an addition to the LD (as the LD corresponds to a residual stream direction, up to layer norm). A model component can easily change the LD by some absolute amount (e.g. +1 LD). It cannot easily change the LD by a relative amount (LD *= 1.5), or change the probabilities by a specific amount (prob += 0.20). For example consider a model component that always outputs -1 logit to duplicated names (assume “John and Mary … John …”). This component then always writes +1 LD in favour of Mary, and gets a score of 1 in terms of LD. Other metrics (such as probability) judge this component differently, depending on what the baseline was (e.g. due to other patches). We would argue that logits and logit differences are closer to the mechanistic process happening in the transformer, and thus feel like a more natural unit. This is of course not a requirement, and also does not hold in all places (e.g. if a component’s output depends on the input LD), but it seems to work well in practice.

\subsection{Flaws \& advantages of different metrics}
It is essential to be aware of what a metric measures and is sensitive to. A key thing to track is whether the metric is discrete vs continuous, and whether it’s exponential vs linear (in the logits) - continuous, linear metrics are usually more accurate, which is crucial when doing exploratory patching and assigning “partial credit” to model components. Here we list common pitfalls of popular metrics.

\begin{figure}[H]
  \centering
  \includegraphics[width=\textwidth]{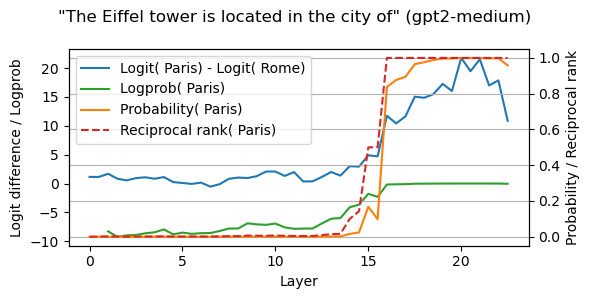}
  \caption{Illustration of different metrics for an example patching experiment with GPT-2 medium.}
  \label{fig}
\end{figure}

\begin{itemize}
    \item Logit difference / logprob difference: The difference between the logit of the correct answer, and the incorrect answer(s). This metric specifically measures the difference between the selected logits, and is not sensitive to components which affect all of them. For example, in IOI it measures the model’s confidence in Mary vs John which encapsulates the IOI-circuit well without being sensitive to the “is the next token a name?”-circuit.
    \item Potential false-positive: Because the metric is a \textit{difference} it may be driven by either getting better at the correct answer or worse at the incorrect answer. Thus it is worth checking the logits or logprobs of individual answers to confirm.
    \\
    This is particularly concerning because the corrupted model likely puts a high probability on the incorrect answer. This means that any patch that indiscriminately damages the model and gets it closer to uniform will damage the incorrect answer logprob and so boost the logit diff.
    \item Logprobs: This metric measures the logprob of the correct answer. It is sensitive to absolute change in logarithmic probabilities (i.e. relative change in probabilities) and captures our intuition for what good model performance means. We broadly think it is a good metric. It’s main flaws are
    \begin{itemize}
        \item Saturation: Once the correct answer becomes the model's top guess, the logprob stops increasing meaningfully, even though the confidence can increase much more.
        \\
        We can see this in Figure \ref{fig}, where the green line saturates after layer 17.
        \item Unspecificity: We lose the ability to control for other properties, e.g. in IOI we cannot distinguish between components that increase both P(John) and P(Mary) from components that only increase P(Mary). This can be intended, or unintended, it’s just important to keep in mind.
        \item Inhibition: To increase the logprob on John, the model can either increase the John logit, or decrease other top logits, and it is hard to distinguish which is happening. This may be desirable or not because the two operations likely have different mechanisms and may be better tracked separately.
    \end{itemize}
    \item Probabilities: This metric measures the probability of the right answer, or the difference in probabilities of different answers. The main issue with such metrics is
    \begin{itemize}
        \item Probabilities are non-linear, in the sense that they track the logits ~exponentially. For example, a model component adding +2 to a given logit can create a 1 or 40 percentage point probability increase, just depending on what the baseline was.
        \\
        As an example of the non-linearity consider the orange line in the figure above: A modest increase in logit difference around layer 17 converts to a jump in probability.
        \item Probabilities also inherit the problems of the logprob metric, namely saturation and unspecificity.
        \\
        The figure shows the saturation effect for the orange line at layer >18.
    \end{itemize}
    \item Binary and discrete metrics (Accuracy / top-k performance / rank / etc): These metrics round off each input to a discrete metric (and then tend to average over a bunch of inputs).
    \begin{itemize}
        \item The problem with these is that generally many components contribute to a model’s performance, with no single decisive contributor. Discrete metrics may suggest that some significant contributors are unimportant, because they aren’t enough to cross a threshold. Alternatively, these metrics may suggest that one contributor among many is \textit{all} that matters because it happens to be the one that  pushes the model over the threshold. We generally recommend using continuous metrics instead.
        \\
        As an example consider the Figure above: The rank-based metric (red line) jumps around layer 15 when the corresponding logit passes the rank 1 and 0 thresholds, while it is not sensitive to any of the other changes.
        \item Discrete metrics can be a good fit for confirmatory patching rather than exploratory patching, as in some sense accuracy \textit{is} the metric we care about - can the model get the question right or not?
    \end{itemize}
    \item Logits: We could just take the answer \textit{logit} as a metric. This is somewhat unprincipled because logits have an arbitrary baseline (adding +1 to all logits would not affect the output) but tend to work in practice. Logit(John) often matches Logprob(John) without being affected by the downsides of the logprob metric.
    \\
    This metric can incorrectly pick up on components that just contribute to many logits. Ensuring that the residual stream and logits have mean zero (default in TransformerLens) can help address this.
\end{itemize}

\begin{figure}[H]
    \centering
    \includegraphics[width=0.9\textwidth]{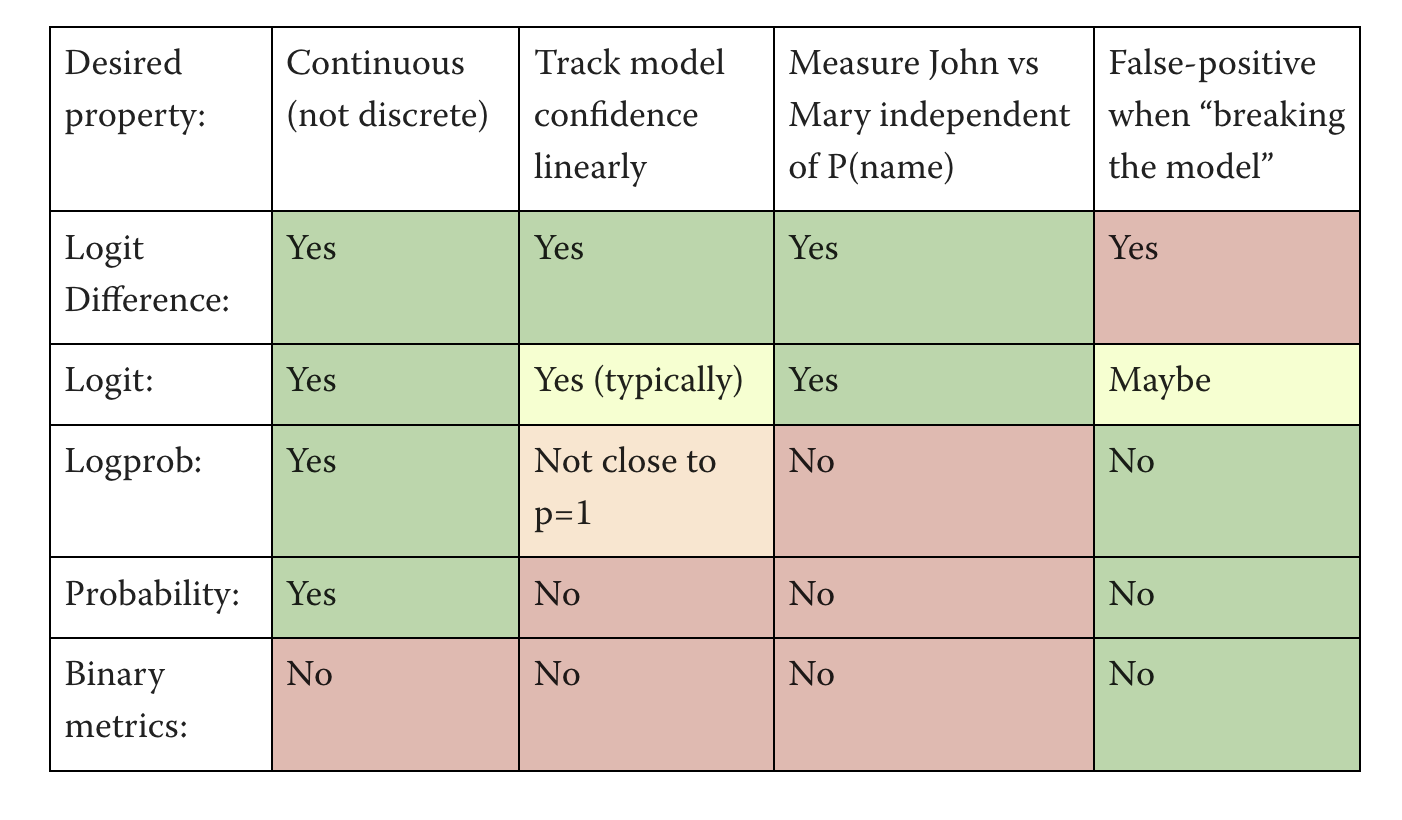}
\end{figure}

\section{Summary}
In most situations, use activation patching instead of ablations. Different corrupted prompts give you different information, be careful about what you choose and try to test a range of prompts.

There are two different directions you can patch in: denoising and noising. These are \textit{not} symmetric. Be aware of what a patching result implies!
\begin{itemize}
    \item Denoising (a clean → corrupt patch) shows whether the patched activations were \textbf{sufficient to restore} the model behaviour. This implies the components make up a cross-section of the circuit.
    \item Noising (a corrupt → clean patch) shows whether the patched activations were \textbf{necessary to maintain} the model behaviour. This implies the components are part of the circuit.
\end{itemize}

Be careful when using metrics that are (i) discrete, (ii) overly sharp, or (iii) sensitive to unintended information. Ideally use a range of metrics, and try to have at least one metric that is continuous and roughly linear in logits such as logit difference or logprob. We recommend representing patching results in a big dataframe with a column per metric and row per patching experiment, and making a bunch of plots from this.
\begin{itemize}
    \item Model top-k accuracy is discrete and can overrepresent changes at thresholds and shows no change for large effects that don't cross thresholds.
    \item Most effects from patching are linear and additive in logit space. Probability is exponential in logit space, so it overemphasises effects near a threshold and suppresses effects elsewhere, creating overly sharp patching plots
    \item Logprob can saturate, and cannot control for a patch that boosts both the correct and incorrect answer(s)
\end{itemize}

\section*{Acknowledgements}
Thanks to Arthur Conmy, Chris Mathwin, James Lucassen, and Fred Zhang for comments on a draft of this manuscript.

\printbibliography

\end{document}